\documentclass{article}
\usepackage{spconf,amsmath,epsfig,amssymb}
\usepackage{graphicx}
\usepackage{multicol,bigstrut,textcomp,bm,subfigure,url}

\let\OLDthebibliography\thebibliography
\renewcommand\thebibliography[1]{
  \OLDthebibliography{#1}
  \setlength{\parskip}{0pt}
  \setlength{\itemsep}{0pt plus 0.3ex}
}

\begin{document}\sloppy

\def\x{{\mathbf x}}
\def\L{{\cal L}}

\title{Deep Convolutional Sparse Coding Network for Pansharpening with Guidance of Side Information }
%
\name{Shuang Xu,
Jiangshe Zhang $^*$ \thanks{*Corresponding author.},
Kai Sun,
Zixiang Zhao,
Lu Huang,
Junmin Liu,
Chunxia Zhang
\thanks{**This work was supported in part by the National Key
	Research and Development Program of China under Grant
	2018AAA0102201, and in part by the National Natural
	Science Foundation of China under Grants 61976174,
	61877049.}}
\address{School of Mathematics and Statistics, Xi'an Jiaotong University, Xi'an 710049, China}

\maketitle

\begin{abstract}
Pansharpening is a fundamental issue in remote sensing field. This paper proposes a side information partially guided convolutional sparse coding (SCSC) model for pansharpening. The key idea is to split the low resolution multispectral image into a panchromatic image related feature map and a panchromatic image irrelated feature map, where the former one is regularized by the side information from panchromatic images. With the principle of algorithm unrolling techniques, the proposed model is generalized as a deep neural network, called as SCSC pansharpening neural network (SCSC-PNN). Compared with 13 classic and state-of-the-art methods on three satellites, the numerical experiments show that SCSC-PNN is superior to others. The codes are available at \url{https://github.com/xsxjtu/SCSC-PNN}.
\end{abstract}
\section{Introduction}
With the launch of satellites, the satellite imagery has played an important role in agriculture, environment and mining. The satellite is usually equipped with two kinds of imaging devices, multispectral (MS) sensors and panchromatic (PAN) sensors. MS images are with high spectral resolution and low spatial resolution, while PAN images are with low spectral resolution and high spatial resolution. To obtain the images with both high spectral and high spatial resolutions, pansharpening that fuses the low resolution MS (LRMS) images and PAN images has received much attention from the remote sensing field. 

Over the past decades, numerous pansharpening algorithms have been developed. The traditional algorithms are classified into two groups, component substitute and multiresolution analysis \cite{BDSD,Brovey,GS,HPF,IHS,Indusion,SFIM}. Nowadays, deep learning has considerable impact on image processing. A growing number of researchers are devoted to improving the pansharpening methods' performance via deep neural networks. The early exploration is the pansharpening neural network (PNN) \cite{PNN}, which is fed with the concatenation of LRMS and PAN images and then regresses the high resolution MS images (HRMS). The simple backbone of PNN, three convolutional layers, greatly limits its performance. Recent researches aim to enhance deep neural networks' performance for the pansharpening task by increasing the number of layers \cite{DRPNN,MSDCNN,RSIFNN,PanNet}. Nonetheless, it is now established that it may dramatically lead to performance degradation if the number of layers exceeds a certain threshold, although this problem can be alleviated by tricks, such as residual learning and dense connection. 

To overcome shortcomings of most deep learning based methods, a research trend is to combine the traditional optimization issues and algorithm unrolling techniques. Very recently, by unrolling a multi-modal convolutional sparse coding (CSC) model, Deng et al. present a novel deep neural network for the general image fusion problem \cite{CUNet}. The key feature is that their network can automatically decompose the images into the common information shared with different modalities, and the unique information of each single modality. Then, in the image fusion stage, it combines both the common information and all unique information to reconstruct the fusion image. In what follows, this network is named as the common and unique information splitting network (CUNet). There is no doubt that CUNet can naturally be adapted to the pansharpening task, but there still remains several problems. For example, CUNet takes the addition strategy to combining the information, so it requires that the source images have the same number of bands. For pansharpening, CUNet has to fuse LRMS and PAN images band-wise, which often ignores the correlation among bands. 

To solve the problems, this paper gets rid of Deng's model and develops a side information partially guided CSC (SCSC) model for pansharpening. Specifically, we formulate a CSC model to extract sparse feature maps of PAN images which will serve as side information. Then, for LRMS images, we formulate two CSC models for LRMS images to separately extract the PAN image related information and the PAN image irrelated information. In addition, the PAN image related information is regularized by the side information from PAN images. In summary, our model incorporates the information from PAN images by a regularization term in the feature space instead of the addition strategy. With the principle of algorithm unrolling techniques, the novel model is generalized as a deep neural network named as SCSC based pansharpening neural network (SCSC-PNN). The experiments conducted on three satellites demonstrate that SCSC-PNN outperforms the classic and the state-of-the-art (SOTA) methods.

\section{Side information guided convolutional sparse coding}
Sparse coding is a technique to obtain representations for signals given a group of atoms from an over-complete dictionary. In the computer vision field, sparse coding can be extended as the convolutional sparse coding (CSC). In formula, the convolutional sparse representation is obtained by solving
\begin{equation}\label{eq:l1}
	\min_{\bm{f}} \left\|\bm{I}-\bm{d}*\bm{f}\right\|_2^2 + \lambda  \left\|\bm{f}\right\|_1,
\end{equation}
where $*$ is convolutional operator, $\bm{I}\in\mathbb{R}^{b\times M\times N}$ is the image with a height of $M$, a width of $N$, and the number of channels $b$, $\bm{d}\in\mathbb{R}^{b\times k\times s\times s}$ is $k$ filters with a size of $s$, and $\bm{f}\in\mathbb{R}^{k\times M\times N}$ is $k$ sparse feature maps. The parameter $\lambda$ controls balance between the data fidelity term and the $\ell_1$-norm sparse penalty. CSC can be further improved if we have prior knowledge of side information on the target image. Given the sparse feature map of the side information $\bm{s}\in\mathbb{R}^{k\times M\times N}$, it is assumed that the $\bm{f}$ not only can reconstruct the target image, but also should be close to $\bm{s}$. To take this prior knowledge into account, the side information guided CSC model \cite{SCSC} can be formulated by inserting an extra penalty into Eq. (\ref{eq:l1}),
\begin{equation}
	\min_{\bm{f}} \left\|\bm{I}-\bm{d}*\bm{f}\right\|_2^2 + \lambda  (\left\|\bm{f}\right\|_1+\left\|\bm{f}-\bm{s}\right\|_1).
\end{equation}

\section{Model formulation}
\subsection{SCSC based pan-sharpening model}
By integrating SCSC model and the idea of algorithm unrolling, we formulate a new network for pan-sharpening. To start with, the PAN image is modeled by the CSC model. Given a PAN image $\bm{P}\in\mathbb{R}^{b\times M\times N}$ (typically $b=1$), it is represented by $\bm{P}=\bm{c}*\bm{z}$, 
where $c\in\mathbb{R}^{1\times k\times s\times s}$ is the filter for PAN images and $\bm{z}\in\mathbb{R}^{k\times s\times s}$ is the corresponding feature map. $\bm{z}$ is inferred by
\begin{equation}\label{eq:model_P}
	\min_{\bm{z}} \left\|\bm{P}-\bm{c}*\bm{z}\right\|_2^2 + \lambda  \left\|\bm{z}\right\|_1.
\end{equation}

As for the LRMS image, we consider its upsampled version $\bm{L}\in \mathbb{R}^{B\times M\times N}$. There is high spatial similarity between $\bm{L}$ and $\bm{P}$, but the spectral response intensity of each band in $\bm{L}$ is also different from $\bm{P}$. In order to describe the relationship between $\bm{L}$ and $\bm{P}$, our model decomposes $\bm{L}$ into a PAN image irrelated part $\bm{a}*\bm{x}$ and a PAN image related part $\bm{b}*\bm{y}$. That is, $\bm{L}=\bm{a}*\bm{x}+\bm{b}*\bm{y}$, where $\bm{a}$ and $\bm{b}$ are the dictionaries for LRMS images. In what follows, $\bm{x}$ is called as the unique feature map, and $\bm{y}$ is called as the common feature map. The prior knowledge from $\bm{P}$ serves as the side information for $\bm{b}*\bm{y}$. In other words, we would like to force $\bm{z}$ and $\bm{y}$ to be closer. Therefore, we formulate the following optimization issue,
\begin{equation}\label{eq:model_LRMS}
\begin{aligned}
	\min_{x,y} & \left\|\bm{L}-(\bm{a}*\bm{x}+\bm{b}*\bm{y})\right\|_2^2 \\
	& + \lambda  (\left\|\bm{x}\right\|_1+\left\|\bm{y}\right\|_1+\left\|\bm{y}-\bm{z}\right\|_1),
\end{aligned}
\end{equation}
where only a part of feature maps (i.e., $\bm{y}$) is guided by side information. 
At last, we assume that LRMS and HRMS images share the feature maps. So, given the dictionary for HRMS images ($\bm{\alpha}$ and $\bm{\beta}$), the HRMS image can be reconstructed by 
\begin{equation}\label{eq:recon}
	\bm{H} = \bm{\alpha}*\bm{x}+\bm{\beta}*\bm{y}.
\end{equation}

In summary, we formulate an SCSC based pansharpening model. The main idea is to model the unique and common feature maps for MS images, where the side information from PAN images regularizes the common feature map. 

\subsection{SCSCPNN}
In our model, it should solve issues (\ref{eq:model_P}) \& (\ref{eq:model_LRMS}) to get the estimations for feature maps $\bm{x}$, $\bm{y}$ and $\bm{z}$. In this section, the solution for each feature map is generalized into a neural module that can be embedded into a network. With the three modules, we present a novel pan-sharpening network. 

\subsubsection{Side information extraction module}
\begin{figure}
	\centering
	\includegraphics[width=1\linewidth]{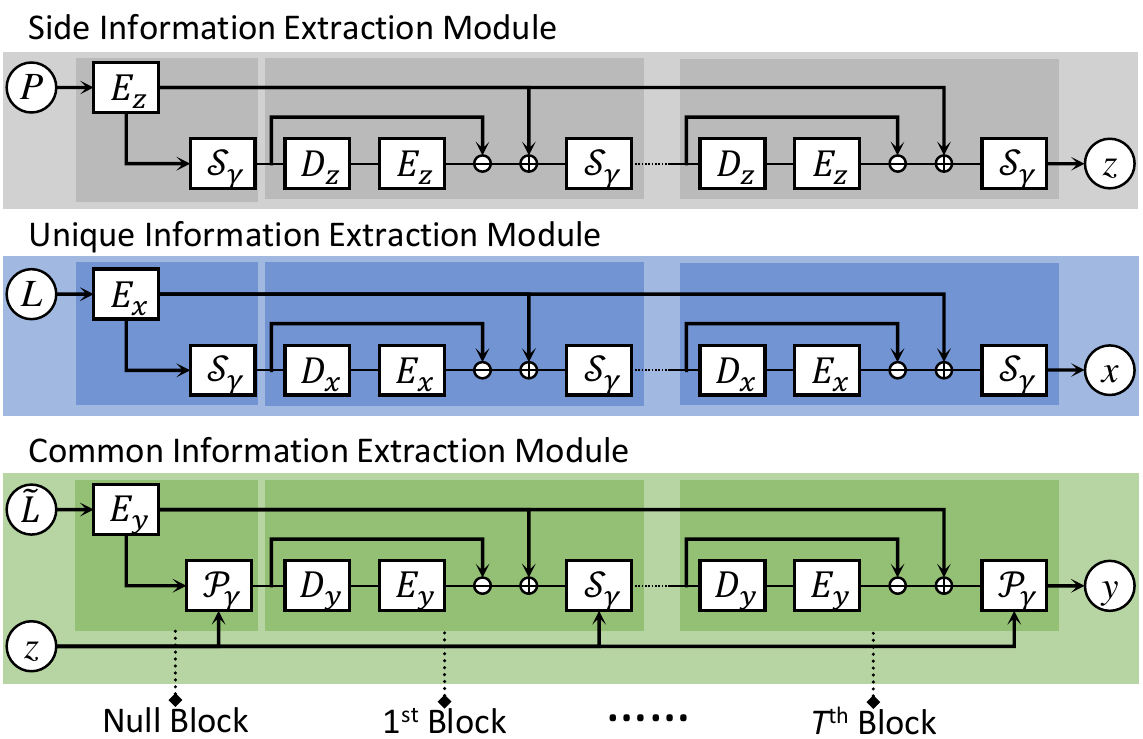}
	\caption{The structures of proposed modules.}
	\label{fig:module}
\end{figure}

To begin with, the solution of Eq. (\ref{eq:model_P}) is converted into a side information extraction module. Eq. (\ref{eq:model_P}) is the $\ell_1$ regularized least squares problem. It can be solved by the iterative shrinkage and thresholding algorithm (ISTA), which applies the regularization's proximal operator to the output of gradient descend algorithm on data fidelity term. In formula, 
\begin{equation}\label{eq:model_P_solution}
	\bm{z}^{(t+1)} = \mathcal{S}_{\gamma} (\bm{z}^{(t)} - \frac{1}{\mu} \bm{c}^T*(\bm{c}*\bm{z}^{(t)}-\bm{P})) ,
\end{equation}
where $\mu$ is a step size, and $\mathcal{S}_{\gamma}(x)=\mathrm{sign}(x)\max(|x|-\gamma,0)$ is the soft thresholding function. Note that $c^T$ is the rotated version of $c$ with 180\textdegree\  along last two dimensions. Recently, the learned ISTA (LISTA) has emerged to translate the iterative step into a neural network. Typically, the filters are replaced by the learnable convolutional layers. According to this principle, the solution of Eq. (\ref{eq:model_P}) can be obtained by a side information extraction module (SIEM), and the computation flow can be written by translating Eq. (\ref{eq:model_P_solution}) as 
\begin{equation}\label{eq:model_P_solution_nn}
	\bm{z}^{(t+1)} = \mathcal{S}_{\bm{\gamma}} (\bm{z}^{(t)} -  \bm{E}_z*\bm{D}_z*\bm{z}^{(t)} + \bm{E}_z*\bm{P}).
\end{equation}
Two learnable convolutional layers $\bm{D}_z$ and $\bm{E}_z$ play the roles of $\bm{c}^T$ and $\bm{c}$. Furthermore, we assign a threshold $\gamma_j(j=1,\cdots,k)$ for each channel of the feature maps and let them learnable. The structure of SIEM is displayed in Fig. \ref{fig:module}. SIEM contains $T$ blocks, each of which carries out the computation flow defined in Eq. (\ref{eq:model_P_solution_nn}). Specially, the feature map $\bm{z}^{(0)}$ is initialized to zero. Thus the computation flow of the null block degenerates as  $\bm{z}^{(1)} = \mathcal{S}_{\gamma} (\bm{E}_z*\bm{P}).$

At last, we analyze the number of parameters in SIEM. Note that $\bm{E}_z\in\mathbb{R}^{k\times b\times s\times s}$, $\bm{D}_z\in\mathbb{R}^{b\times k\times s\times s}$ and $\bm{\gamma}=(\gamma_1,\cdots,\gamma_k)\in\mathbb{R}^{k}$. For each recurrent block, there are an $\bm{E}_z$, a $\bm{D}_z$ and an $\mathcal{S}_{\bm{\gamma}}(\cdot)$, and thus there are $2bks^2+k$ parameters. For the null block, there is no $\bm{D}_z$, so there are $bks^2+k$ parameters. In summary, an SIEM with $T$ blocks has $(2T+1)bks^2+(T+1)k$ parameters.

\begin{figure}
	\centering
	\includegraphics[width=1\linewidth]{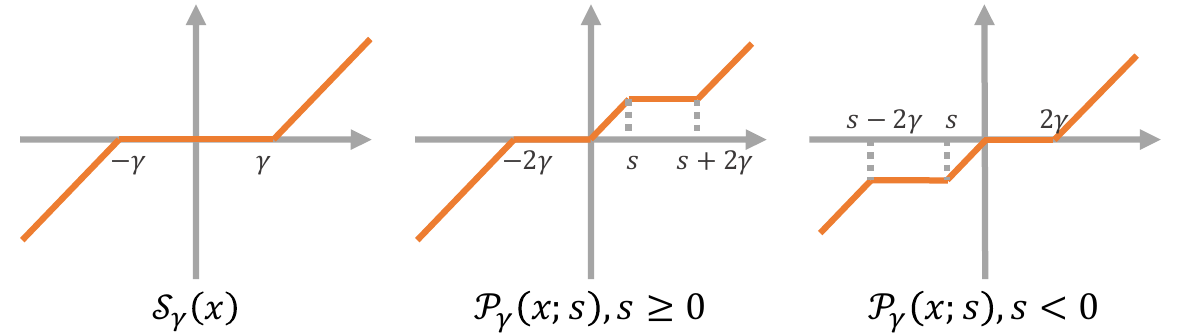}
	\caption{(left) The soft thresholding function. (middle\&right) The piece-wise soft thresholding function.}
	\label{fig:sst}
\end{figure}
\subsubsection{Unique information extraction module}
There are two unknown terms in Eq. (\ref{eq:model_LRMS}). Firstly, we fix $\bm{y}$ and update $\bm{x}$. Now, Eq. (\ref{eq:model_LRMS}) is cast as the following issue,
\begin{equation}\label{eq:model_X}
\min_{\bm{x}}  \|\hat{\bm{L}}-\bm{a}*\bm{x}\|_2^2 + \lambda  \left\|\bm{x}\right\|_1,
\end{equation}
where $\hat{\bm{L}}=\bm{L}-\bm{b}*\bm{y}$. The ISTA's iterative step is
\begin{equation}\label{eq:model_X_solution}
\bm{x}^{(t+1)} = \mathcal{S}_{\gamma} (\bm{x}^{(t)} - \frac{1}{\mu} \bm{a}^T*(\bm{a}*\bm{x}^{(t)}-\hat{\bm{L}})) ,
\end{equation}
Similar to Eqs. (\ref{eq:model_P}) \& (\ref{eq:model_P_solution_nn}), the solution of issue (\ref{eq:model_X}) is also translated by a unique information extraction module (UIEM), i.e.,
\begin{equation}
	\bm{x}^{(t+1)} = \mathcal{S}_{\bm{\gamma}} (\bm{x}^{(t)} -  \bm{E}_x*\bm{D}_x*\bm{x}^{(t)} + \bm{E}_x*\hat{\bm{L}}).
\end{equation}
The structure of UIEM is displayed in Fig. \ref{fig:module}. Similar to SIEM, the number of parameters in UIEM is $(2T+1)Bks^2+(T+1)k$. The inputs of the UIEM reply on $\hat{\bm{L}}=\bm{L}-\bm{b}*\bm{y}$. Nonetheless, in our network, the common feature $\bm{y}$ is computed via a common information extraction module after UIEM. We thus initialize $\bm{y}=\bm{0}$ for UIEM. In this manner, the input of the UIEM becomes $\hat{\bm{L}}=\bm{L}$. 

\begin{figure}[]
	\centering
	\includegraphics[width=1\linewidth]{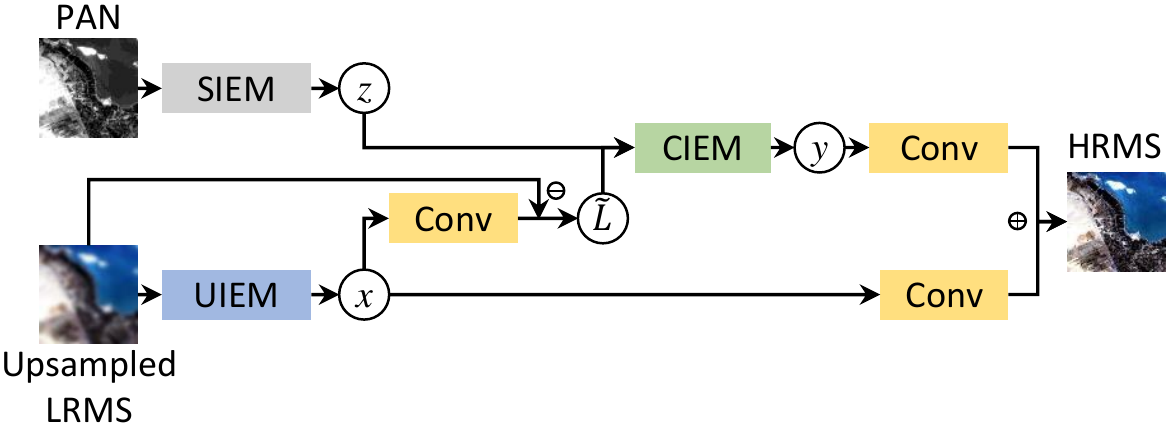}
	\caption{The architecture of proposed SCSC-PNN.}
	\label{fig:net}
\end{figure}

\begin{figure*}[t]
	\centering
	\subfigure[ {\scriptsize LRMS}] {\includegraphics[width=.15\linewidth]{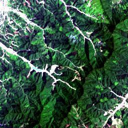}}  
	\subfigure[ {\scriptsize PAN}] {\includegraphics[width=.15\linewidth]{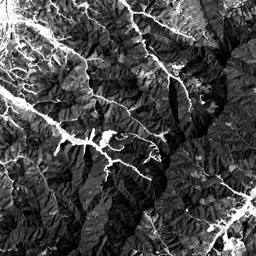}}  
	\subfigure[ {\scriptsize GT}] {\includegraphics[width=.15\linewidth]{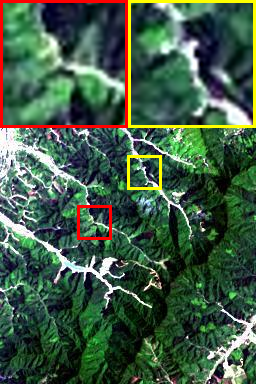}}  
	\subfigure[ {\scriptsize BDSD}] {\includegraphics[width=.15\linewidth]{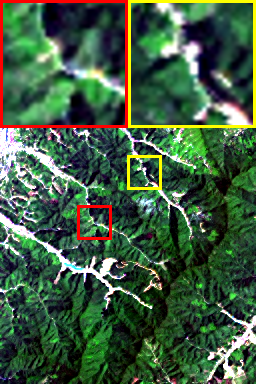}}    
	\subfigure[ {\scriptsize GS}] {\includegraphics[width=.15\linewidth]{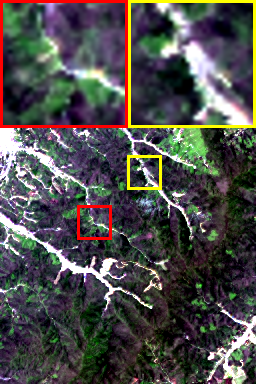}} 
	\subfigure[ {\scriptsize MIPSM}] {\includegraphics[width=.15\linewidth]{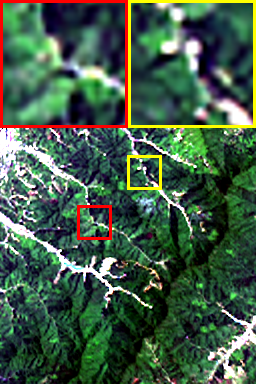}} 
	\subfigure[ {\scriptsize DRPNN}] {\includegraphics[width=.15\linewidth]{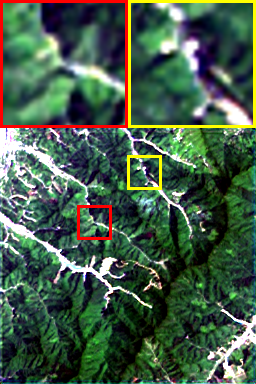}} 
	\subfigure[ {\scriptsize MSDCNN}] {\includegraphics[width=.15\linewidth]{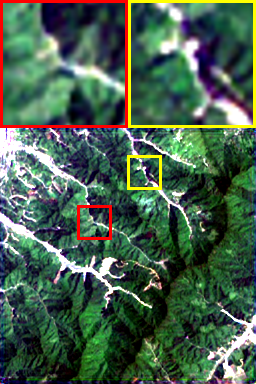}} 
	\subfigure[ {\scriptsize RSIFNN}] {\includegraphics[width=.15\linewidth]{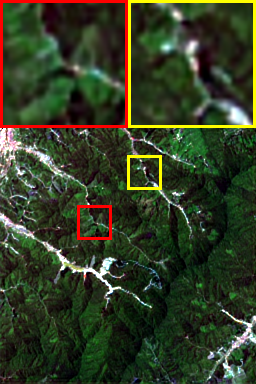}} 
	\subfigure[ {\scriptsize PANNET}] {\includegraphics[width=.15\linewidth]{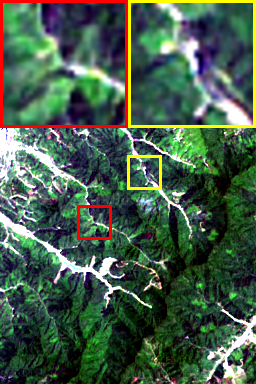}} 
	\subfigure[ {\scriptsize CUNet}] {\includegraphics[width=.15\linewidth]{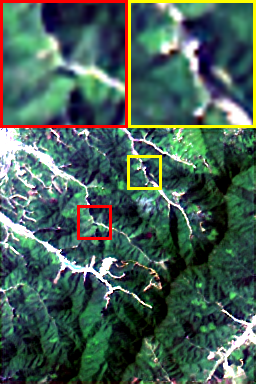}} 
	\subfigure[ {\scriptsize SCSCPNN}] {\includegraphics[width=.15\linewidth]{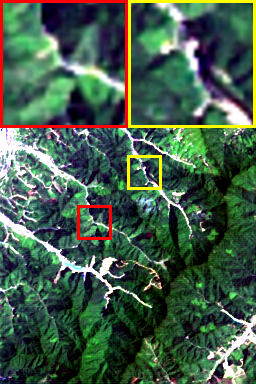}}    
	\caption{Visual inspection on Landsat8 dataset.}
	\label{fig:L8}
\end{figure*}

\begin{figure*}[h]
	\centering
	\subfigure[ {\scriptsize LRMS}] {\includegraphics[width=.15\linewidth]{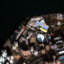}}  
	\subfigure[ {\scriptsize PAN}] {\includegraphics[width=.15\linewidth]{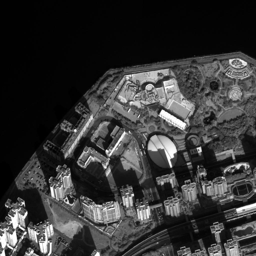}}  
	\subfigure[ {\scriptsize GT}] {\includegraphics[width=.15\linewidth]{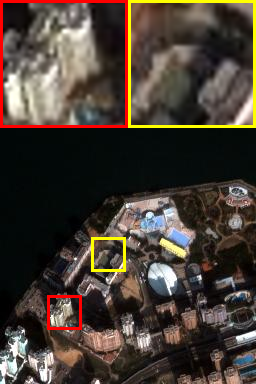}}  
	\subfigure[ {\scriptsize BDSD}] {\includegraphics[width=.15\linewidth]{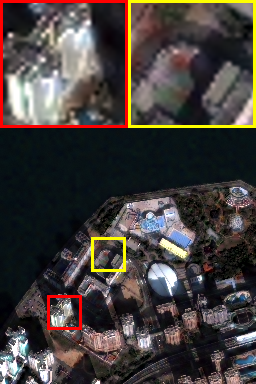}}    
	\subfigure[ {\scriptsize GS}] {\includegraphics[width=.15\linewidth]{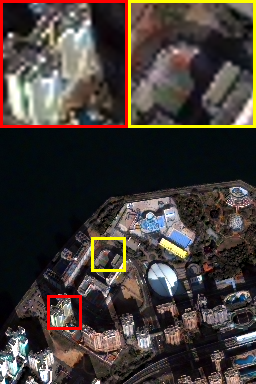}} 
	\subfigure[ {\scriptsize MIPSM}] {\includegraphics[width=.15\linewidth]{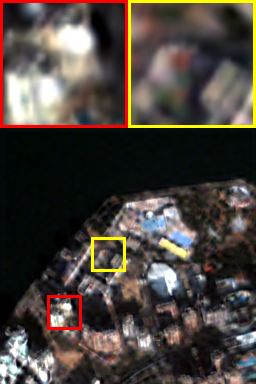}} 
	\subfigure[ {\scriptsize DRPNN}] {\includegraphics[width=.15\linewidth]{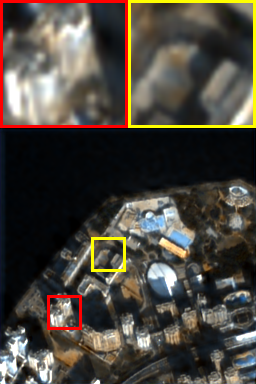}} 
	\subfigure[ {\scriptsize MSDCNN}] {\includegraphics[width=.15\linewidth]{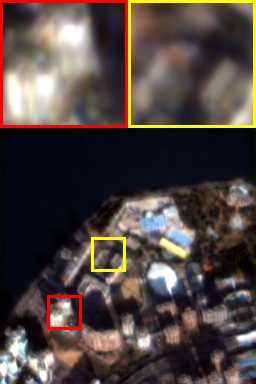}} 
	\subfigure[ {\scriptsize RSIFNN}] {\includegraphics[width=.15\linewidth]{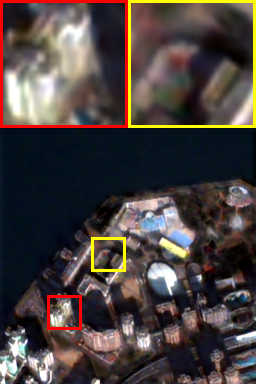}} 
	\subfigure[ {\scriptsize PANNET}] {\includegraphics[width=.15\linewidth]{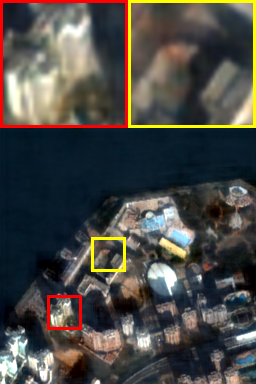}} 
	\subfigure[ {\scriptsize CUNet}] {\includegraphics[width=.15\linewidth]{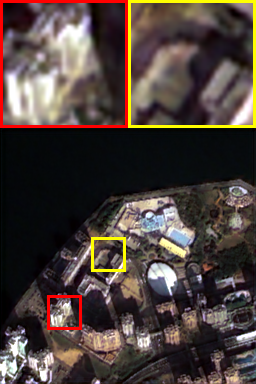}} 
	\subfigure[ {\scriptsize SCSCPNN}] {\includegraphics[width=.15\linewidth]{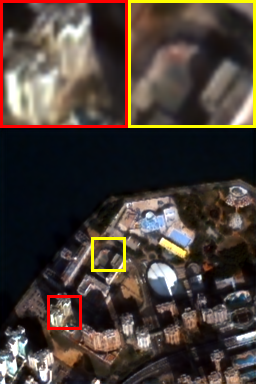}}    
	\caption{Visual inspection on QuickBird dataset.}
	\label{fig:QB}
\end{figure*}

\begin{figure*}[h]
	\centering
	\subfigure[ {\scriptsize Landsat8}] {\includegraphics[width=.30\linewidth]{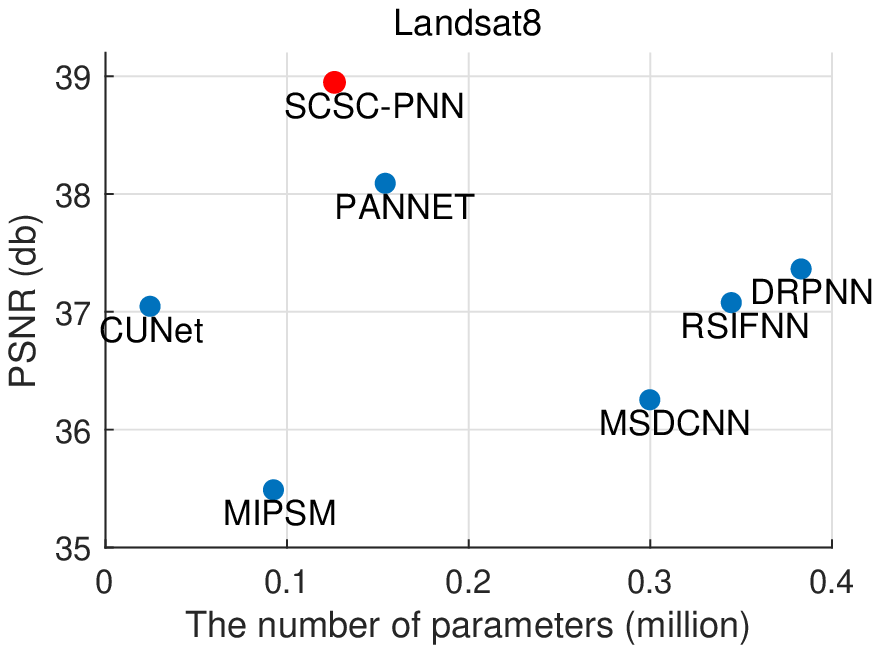}}  
	\subfigure[ {\scriptsize QuickBird}] {\includegraphics[width=.30\linewidth]{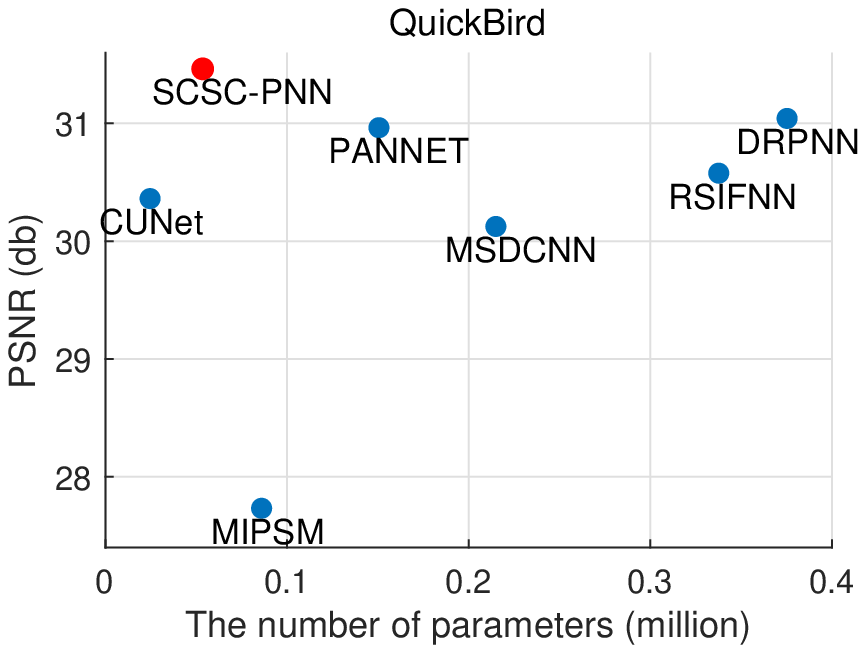}}  
	\subfigure[ {\scriptsize GaoFen2}] {\includegraphics[width=.30\linewidth]{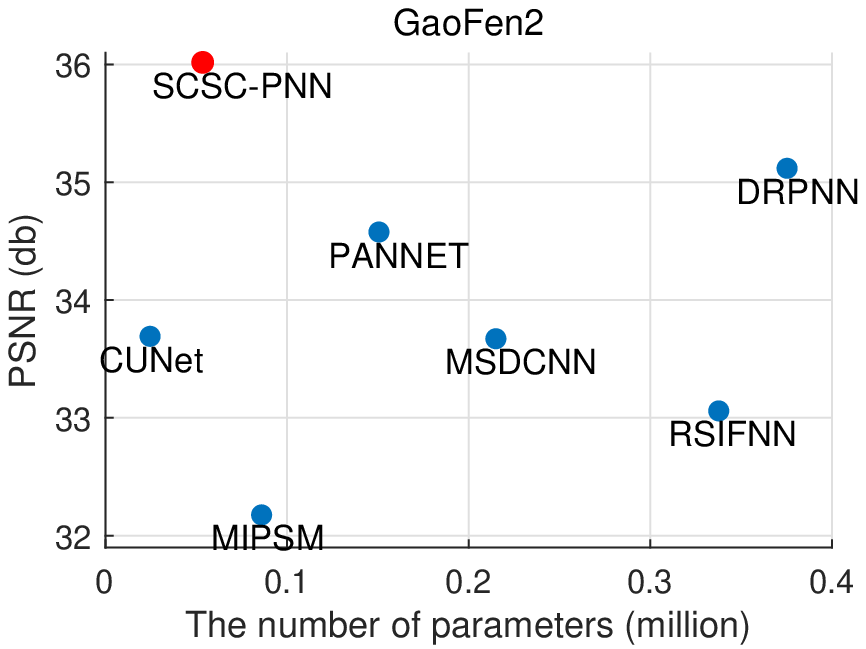}}  
	\caption{PSNR vs. the number of parameters. The parameter of networks depends on the number of bands in the MS images.}
	\label{fig:Param}
\end{figure*} 

\subsubsection{Common information extraction module}
At last, given the side information $\bm{z}$ and unique information $\bm{x}$, it is ready to derive the common information extraction module (CIEM). With regard to $\bm{y}$, Eq. (\ref{eq:model_LRMS}) is cast as the following issue,
\begin{equation}
	\min_{\bm{y}}  \|\tilde{\bm{L}}-\bm{b}*\bm{y}\|_2^2 
	 + \lambda  (\left\|\bm{y}\right\|_1+\left\|\bm{y}-\bm{z}\right\|_1),
\end{equation}
where $\tilde{\bm{L}}=\bm{L}-\bm{a}*\bm{x}$. The iterative step of this issue is 
\begin{equation}\label{eq:model_Y_solution}
\bm{y}^{(t+1)} = \mathcal{P}_{\gamma} (\bm{y}^{(t)} - \frac{1}{\mu} \bm{b}^T*(\bm{b}*\bm{y}^{(t)}-\tilde{\bm{L}});\bm{z}) .
\end{equation}
Here, $\mathcal{P}_{\gamma}(x;s)$ is the piece-wise soft thresholding function, and it is defined by

1. if $s \geq 0$,
\begin{equation}
\mathcal{P}_{\gamma}(x;s)=\left\{\begin{array}{lr}
x+2 \gamma, & x<-2 \gamma \\
0, & -2 \gamma \leq x \leq 0 \\
x, & 0<x<s \\
s, & s \leq x \leq s+2 \gamma \\
x-2 \gamma, & x \geq s+2 \gamma
\end{array}\right. ,
\end{equation}

2. if $s<0$,
\begin{equation}
\mathcal{P}_{\gamma}(x;s)=\left\{\begin{array}{lr}
x+2 \gamma, & x<s-2 \gamma \\
s, & s-2 \gamma \leq x \leq s \\
x, & s<x<0 \\
0, & 0 \leq x \leq 2 \gamma \\
x-2 \gamma, & x \geq 2 \gamma
\end{array}\right. .
\end{equation}
With the similar techniques, it is able to write the computation flow of the CIEM, namely,
\begin{equation}
	\bm{y}^{(t+1)} = \mathcal{P}_{\bm{\gamma}} (\bm{y}^{(t)} -  \bm{E}_y*\bm{D}_y*\bm{y}^{(t)} + \bm{E}_y*\tilde{\bm{L}}).
\end{equation}
The structure of CIEM is displayed in Fig. \ref{fig:module}. The number of parameters in CIEM is $(2T+1)Bks^2+(T+1)k$. CIEM infers the sparse feature maps from the residual image $\tilde{\bm{L}}=\bm{L}-\bm{a}*\bm{x}$. To make our network flexible, when we project $\bm{x}$ from the feature space into the image space, the filter $\bm{a}$ is replaced by a learnable convolutional layer. 

\subsubsection{Pan-sharpening neural network}
The architecture of our proposed SCSC-PNN is displayed in Fig. \ref{fig:net}. Firstly, the PAN/LRMS image passes through SIEM/UIEM to infer $\bm{z}$/$\bm{x}$. Secondly, a convolutional unit projects $\bm{x}$ from the feature space into the image space, and we compute the residual image, $\tilde{\bm{L}}$. Thirdly, CIEM extracts the common feature $\bm{y}$ from residual image $\tilde{\bm{L}}$ with the guidance of side information $\bm{z}$. At last, according to Eq. (\ref{eq:recon}), we replace the dictionary $\bm{\alpha}$ and $\bm{\beta}$ by two convolutional units in order to map $\bm{x}$ and $\bm{y}$ into the image space, and reconstruct the HRMS $\hat{\bm{H}}$. In summary, SCSC-PNN contains an SIEM, a UIEM, a CIEM and three convolutional layers. There are $(2T+1)(b+2B)ks^2+3(T+1)k+3kBs^2$ parameters. 

SCSC-PNN is supervised by the $\ell_1$ loss between the ground truth $\bm{H}$ and the reconstruction $\hat{\bm{H}}$, namely,
$
	Loss = \|\bm{H}-\hat{\bm{H}}\|_1.
$
Throughout our experiments, the configuration of SCSC-Net is set as follows: kernel size $s=3$, the number of filters $k=64$ and the number of blocks $T=4$. SCSC-Net is optimized by Adam over 200 epochs with a learning rate of $5\times10^{-4}$ and a batch size of 8.

\section{Experiments}
SCSC-PNN is compared with six state-of-the-art (SOTA) methods, namely, 
MIPSM~\cite{MIPSM}, 
DRPNN~\cite{DRPNN}, 
MSDCNN~\cite{MSDCNN}, 
RSIFNN~\cite{RSIFNN}, 
PanNet~\cite{PanNet} 
and CUNet~\cite{CUNet}. 
Our method is also compared with seven classic methods, including 
BDSD method \cite{BDSD}, 
Brovey~\cite{Brovey}, 
GS~\cite{GS}, 
HPF~\cite{HPF}, 
IHS~\cite{IHS}, 
Indusion~\cite{Indusion}, 
SFIM~\cite{SFIM}. The experiments are conducted on a computer with an Intel i7-9700K CPU at 3.60GHz and an NVIDIA GeForce RTX 2080ti GPU. 

\subsection{Datasets and metrics}
Three satellites, Landsat8, QuickBird and GaoFen2, are selected to construct the datasets. On Landsat8 and GaoFen2, there are 350/50/100 images for training/validation/test. On QuickBird, there are 474/103/100 images for training/validation/test. The MS images capture by Landsat8 have 10 bands and the spatial up-scaling ratio is 2. For QuickBird and GaoFen2, their MS images have 4 bands and the spatial up-scaling ratio is 4. In the training phase, the MS images are cropped into $32\times32$ patches, and the PAN images are cropped into corresponding patches. Furthermore, Wald protocol is exploited to generate the training samples. In the test phase, we employ three spatial assessment metrics (PSNR, SSIM, ERGAS) and a spectral assessment metric (SAM) to evaluate the methods' performances.

\begin{table*}[]
	\centering
	\caption{Metrics on three test datasets. The best and the second best are highlighted by \textbf{bold} and \underline{underline}, respectively. The up or down arrow indicates higher or lower metric corresponds to better images.}
	\resizebox{\textwidth}{!}{
		\begin{tabular}{|l|cccc|cccc|cccc|}
			\hline
			& \multicolumn{4}{c|}{Landsat8} & \multicolumn{4}{c|}{QuickBird} & \multicolumn{4}{c|}{GaoFen2} \bigstrut\\
			\cline{2-13}          & PSNR$\uparrow$ & SSIM$\uparrow$ & SAM$\downarrow$ & ERGAS$\downarrow$ & PSNR$\uparrow$ & SSIM$\uparrow$ & SAM$\downarrow$ & ERGAS$\downarrow$ & PSNR$\uparrow$ & SSIM$\uparrow$ & SAM$\downarrow$ & ERGAS$\downarrow$ \bigstrut\\
			\hline
			BDSD  & 33.8065  & 0.9128  & 0.0255  & 1.9128  & 23.5540  & 0.7156  & 0.0765  & 4.8874  & 30.2114  & 0.8732  & 0.0126  & 2.3963  \bigstrut[t]\\
			Brovey & 32.4030  & 0.8533  & 0.0206  & 1.9806  & 25.2744  & 0.7370  & 0.0640  & 4.2085  & 31.5901  & 0.9033  & 0.0110  & 2.2088  \\
			GS    & 32.0163  & 0.8687  & 0.0304  & 2.2119  & 26.0305  & 0.6829  & 0.0586  & 3.9498  & 30.4357  & 0.8836  & 0.0101  & 2.3075  \\
			HPF   & 32.6691  & 0.8712  & 0.0250  & 2.0669  & 25.9977  & 0.7378  & 0.0588  & 3.9452  & 30.4812  & 0.8848  & 0.0113  & 2.3311  \\
			IHS  & 32.8772  & 0.8615  & 0.0245  & 2.3128  & 24.3826  & 0.6742  & 0.0647  & 4.6208  & 30.4754  & 0.8639  & 0.0108  & 2.3546  \\
			Indusion & 30.8476  & 0.8168  & 0.0359  & 2.4216  & 25.7623  & 0.6377  & 0.0674  & 4.2514  & 30.5359  & 0.8849  & 0.0113  & 2.3457  \\
			SFIM  & 32.7207  & 0.8714  & 0.0248  & 2.0775  & 24.0351  & 0.6409  & 0.0739  & 4.8282  & 30.4021  & 0.8501  & 0.0129  & 2.3688  \\
			MIPSM & 35.4891  & 0.9389  & 0.0209  & 1.5769  & 27.7323  & 0.8411  & 0.0522  & 3.1550  & 32.1761  & 0.9392  & 0.0104  & 1.8830  \\
			DRPNN & 37.3639  & 0.9613  & 0.0173  & 1.3303  & \underline{31.0415} & \underline{0.8993} & 0.0378  & \underline{2.2250} & \underline{35.1182} & 0.9663  & 0.0098  & \underline{1.3078} \\
			MSDCNN & 36.2536  & 0.9581  & 0.0176  & 1.4160  & 30.1245  & 0.8728  & 0.0434  & 2.5649  & 33.6715  & \underline{0.9685} & 0.0090  & 1.4720  \\
			RSIFNN & 37.0782  & 0.9547  & 0.0172  & 1.3273  & 30.5769  & 0.8898  & 0.0405  & 2.3530  & 33.0588  & 0.9588  & 0.0112  & 1.5658  \\
			PANNET & \underline{38.0910} & \underline{0.9647} & \underline{0.0152} & \underline{1.3021} & 30.9631  & 0.8988  & \underline{0.0368} & 2.2648  & 34.5774  & 0.9635  & \underline{0.0089} & 1.4750  \\
			CUNet & 37.0468 & 0.9610 & 0.0179 & 1.3430 & 30.3612 & 0.8876 & 0.0428 & 2.4178 & 33.6919 & 0.9630 & 0.0184 & 1.5839 \\
			SCSC-PNN & \textbf{39.3156} & \textbf{0.9730} & \textbf{0.0129} & \textbf{1.1849} & \textbf{31.4776} & \textbf{0.9058} & \textbf{0.0352} & \textbf{2.0899} & \textbf{36.1019} & \textbf{0.9735} & \textbf{0.0083} & \textbf{1.1960} \bigstrut[b]\\
			\hline
		\end{tabular}%
	}
	\label{tab:result}%
\end{table*}%

\subsection{Comparison with SOTA methods}
The results are reported in Table \ref{tab:result}. It is shown that SCSC-PNN outperforms others on all datasets with regard to all metrics. SCSC-PNN increases the PSNR values by 0.85db, 0.42db and 0.90db on Landsat8, QuickBird and GaoFen2, respectively. Furthermore, the scatter plot of PSNR vs. the number of parameters shown in Fig. \ref{fig:Param} demonstrates that our improvement benefits from the network structure instead of the number of parameters, because SCSC-PNN has fewer parameters but better performance. From the pansharpened images of representative methods shown in Figs. \ref{fig:L8} and \ref{fig:QB}, it is learned that our images are closest to the ground truth and have less spatial as well as spectral distortions. 

\subsection{Ablation experiments}
The number of blocks $T$ plays an important role. We evaluate performance of SCSC-PNN with different $T$ on Landsat8. As shown in Table \ref{tab:T}, SCSC-PNN with larger $T$ tends to perform better, but the improvement is not obvious when $T>4$. To strike the balance of performance and parameters, $T=4$ is a good choice.

\begin{table}[htbp]
	\centering
	\caption{Performance with different $T$ on validation set of Landsat8.}
		\resizebox{\linewidth}{!}{
	\begin{tabular}{|c|cccccc|}
		\hline
		$T$ & 1     & 2     & 3     & 4     & 5     & 6  \bigstrut\\
		\hline
		PSNR$\uparrow$ & 38.4638  & 38.5338  & 38.7148  & 39.0488  & 38.9013  & 39.0334  \bigstrut[t]\\
		SSIM$\uparrow$ & 0.9677  & 0.9690  & 0.9699  & 0.9714  & 0.9708  & 0.9723  \\
		SAM$\downarrow$ & 0.0153  & 0.0155  & 0.0144  & 0.0141  & 0.0144  & 0.0138  \\
		ERGAS$\downarrow$ & 1.2383  & 1.2012  & 1.2692  & 1.1562  & 1.1976  & 1.1899  \bigstrut[b]\\
		\hline
	\end{tabular}%
}
	\label{tab:T}%
\end{table}%

\section{Conclusion}
In this paper, we propose a model based pansharpening network. Considering that MS images contains the PAN related and irrelated features, we formulate a SCSC model to recover HRMS images, which is then generalized into a deep networks called by SCSC-PNN. The extensive experiments show that SCSC-PNN outperforms other counterparts and has fewer parameters.

\bibliographystyle{IEEEbib}
\bibliography{icme2021template}

\end{document}